\documentclass[runningheads]{llncs}

\usepackage{amssymb}
\usepackage{graphicx}

\usepackage{url}

\usepackage{booktabs}
\usepackage{array}
\usepackage{graphicx}
\usepackage{amsmath}
\usepackage{multirow}

\def\footnoterule{\relax%
  \kern-5pt
  \hbox to \columnwidth{\hfill\vrule width 0.5\columnwidth height 0.4pt\hfill}
  \kern4.6pt}
\makeatother

\urldef{\mailal}\path|alan.smeaton@dcu.ie|    
\newcommand{\keywords}[1]{\par\addvspace\baselineskip
\noindent\keywordname\enspace\ignorespaces#1}

\begin{document}

\mainmatter  

\title{Using GANs to Synthesise  Minimum Training Data for Deepfake Generation}


%
%
\author{Simranjeet Singh\textsuperscript{1} \and Rajneesh Sharma\textsuperscript{1} \and Alan F. Smeaton\textsuperscript{1,2}}
\authorrunning{Singh et al.}

\institute{\textsuperscript{1}School of Computing and
\textsuperscript{2}Insight Centre for Data Analytics\\
Dublin City University, Glasnevin, Dublin 9, Ireland\\
\mailal\\
}

%
%

\toctitle{Using GANs to Synthesise  Minimum Training Data for Deepfake Generation}
\tocauthor{Authors' Instructions}
\maketitle

\begin{abstract}
There are many applications of Generative Adversarial Networks 
(GANs)  in fields like computer vision, natural language processing, speech synthesis, and more.  Undoubtedly the most notable results have been in the area of image synthesis and in particular in the generation of deepfake videos.  While deepfakes have received much negative media coverage, they can be a useful technology in applications like entertainment, customer relations, or even assistive care. One problem with generating deepfakes is the requirement for a lot of image training data of the subject which is not an issue if the subject is a celebrity for whom many images already exist. If there are only a small number of training images then the quality of the deepfake will be poor.  Some media reports have indicated that a good deepfake can be produced with as few as 500 images but in practice, quality deepfakes require many thousands of images, one of the reasons why deepfakes of celebrities and politicians have become so popular.  In this study, we exploit the property of a GAN to produce images of an individual with variable facial expressions which we then use to generate a deepfake.  We observe that with such variability in facial expressions of synthetic GAN-generated training images and a  reduced quantity of them, we can produce a near-realistic deepfake videos.

\keywords{Deepfake generation; Generative Adversarial Networks; GANs; Variable face images.}
\end{abstract}

\section{Introduction}

Recently we have seen a rise in the presence of deepfake videos on social media and in entertainment applications. Sometimes these are used for good but it is the mis-use of deepfakes which attracts most media attention and commentary.  What makes deepfakes so important today is their low barrier to entry, meaning that easily available tools and models can be used by researchers with even moderate programming skills to generate very realistic deepfake videos.  When this is considered in the context of targeted advertisements for political elections on social media, then the impact of deepfakes could be quite significant.

A deepfake is a video created by manipulating an original video using advanced machine learning techniques. This involves replacing the face of an individual from a source video with the face of a second person in the destination video.  A model of the face of the second person, the one who is superimposed into the destination video, is created based on a typically large collection of facial images. In the early days of deepfake videos, celebrities were used in the destination videos because (a) it is easy to get thousands of images of celebrities from the internet and (b) most of these pictures are of the subject facing the camera. The Hollywood actor Nicholas Cage became even more of a celebrity as a model based on images of his face was one of the first to be made publicly available and was widely used in creating deepfakes when the interest was in the quality of the generated videos and less on who the subjects were.

Now that we have reached the point where the quality of deepfakes is almost indiscernible from real videos, interest  returns to how to generate these deepfakes, not using celebrities as the subjects but using  ordinary people.  
While there are nefarious applications based on the use of deepfakes of non-celebrity individuals, there are also useful scenarios. An example of this is using deepfake videos of a non-celebrity as a sales agent or troubleshooter in an online chat system.

One characteristic of the non-celebrity subject in a deepfake, is that there will typically be a limited number of images of the subject’s face available for training a deepfake generator, perhaps even no images to start from. Thus we expect that training data, i.e.  images of the face, may actually be taken from short video clips recorded specifically for this purpose.  

In this paper we look at how deepfake videos of non-celebrity subjects can be generated using limited training data, i.e. a small number of training images.  In particular we are interested not just in the limited number of images used but also in the variability of facial expressions among those limited number of images.  To test this we use a large number of images to create a model of an individual face, and then we generate a small number of synthetic but realistic images from that model which we use to generate a deepfake.  While this may seem counter intuitive, to use a large number of images of a celebrity to generate a small number of synthetic images of that celebrity this  allows the synthetic images to include a lot of facial variety of expression which we could not obtain easily if we were to use a real collection as the small number of deepfake training images.

The rest of this paper is organised as follows. In the next section we present an overview of Generative Adversarial Networks (GANs) followed by a description of 4 metrics used to evaluate the quality of our output from image-generating GANs. We then describe how we gathered, or more correctly how we synthesised image data for training a GAN and we then present an analysis of those images in terms of their quality and variability of facial expressions. That is followed by a description of how we used those images to create a deepfake and then some conclusions and plans for future work.

\section{Generative Adversarial Networks (GANs)}

The idea behind adversarial networks was first published by Olli Niemitalo however his ideas were never implemented \cite{2_rose_2014} and  a similar concept was introduced by Li, Gauci and Gross in 2013 \cite{2_rose_2014}. Generative Adversarial Network (GAN) implementations were first described in 2014 by Ian Goodfellow and until 2017 the use of GANs was restricted to just image enhancement  to produce high quality images. In 2017 GANs were used for the first time for generating new facial images and the idea began to make its presence known in the fine arts arena and were thus dubbed  creative adversarial networks \cite{2_rose_2014}.

GANs have been widely applied to  domains such as computer vision, natural language processing, etc. GANs have contributed immensely to the field of image generation \cite{karras2019analyzing} where the quality of synthetic images a GAN can produce has improved significantly over the years since its inception. 
Other example applications of GANs include the generation of DNA sequences, 3D models of replacement teeth, impressionist paintings, and of course video clips, some known as deepfakes.

Deepfakes are a form of video manipulation where two trained networks are pitted against each other to generate an output of sufficient quality as to be close to indecipherable. They operate by inputting a set of images of a subject from which they build a model of the face and then superimpose this face model on the target face in an original video. 

One of the challenges faced by deepfake generation, apart from their computational cost, is the requirement for a large number of training images of the subject to be faked into the original image. In practice, the quality of the generated deepfake will depend not only on the number of face images in the training data but the amount of facial variability among those images and the amount of facial variation in the original video. If the original video has a  face with not much emotion shown and very little variation in facial expression then it follows that the training data for the face to be superimposed does not need a wide variety of facial expression and thus a smaller number of training images are needed. If the original video has a lot of facial variation then the model to be generated to replace this original face will need to be larger and more complex, and thus require far more training data.  Some commentators have said that as few as 500 images of the face of a subject are required for a good deepfake but in practice these refer to deepfakes without much facial emotion and the best deepfakes are generated using many thousands of source images of the face. 

Deepfakes have many applications in the entertainment industry such as movie production and the Salvador Dali museum in St Petersburg, Florida\footnote{\url{https://thedali.org/}}, but there are also applications in areas like customer relations where text or audio chatbots are replaced by real people or deepfakes, or in assistive technology where older people living alone might interact with generated media which could consist of deepfaked videos of loved ones.  The problem with such applications is that there are usually few images available from which to train a model to create a deepfake.

\begin{figure}[ht]
    \centering
    \includegraphics[width=\columnwidth]{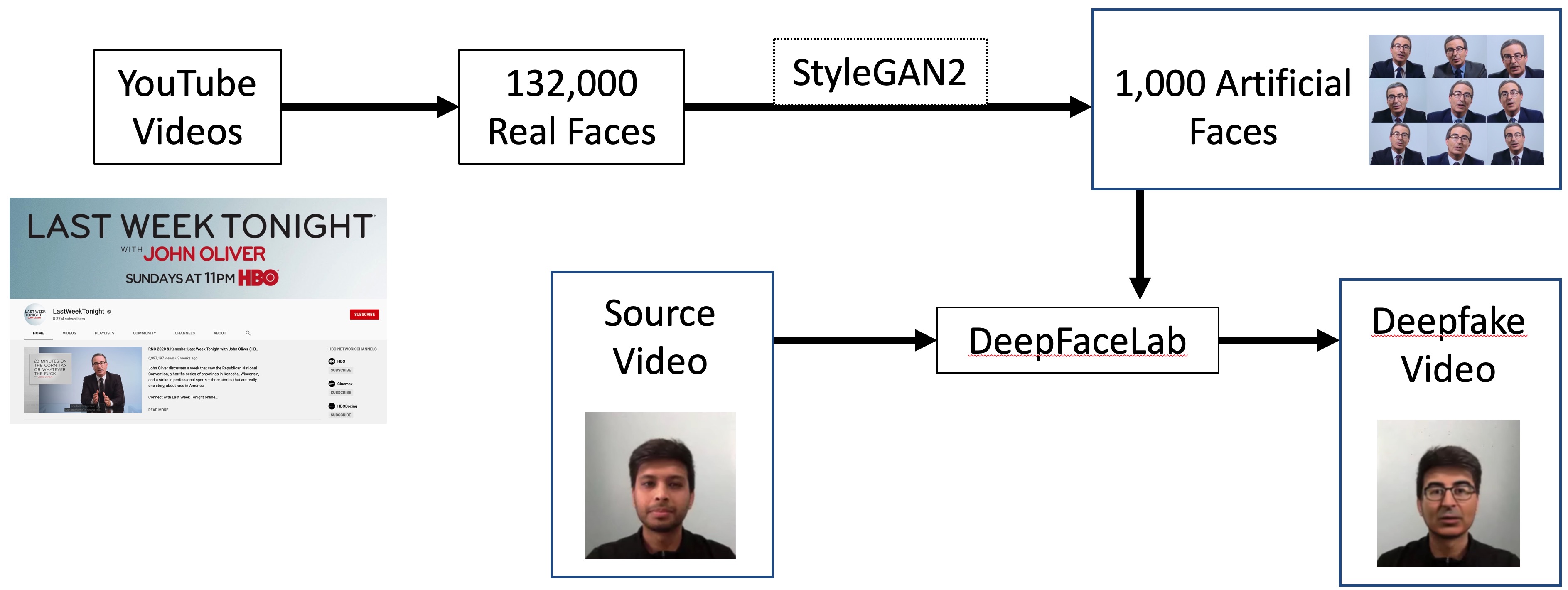}
    \caption{Overview of data processing in the paper}
    \label{fig:System}
\end{figure}

In this study we look into how the amount of,  and the variety of facial expressions included in, the face data used to train a deepfake generator affects the quality of the deepfake. One of the latest GANs, StyleGAN2 \cite{karras2019analyzing}, is used in our study to produce synthetic facial images for training and various evaluation methods are  used to benchmark the quality  of these synthetic images  including Inception score \cite{salimans2016improved} and the Fr{\'e}chet Inception Distance \cite{heusel2017gans}, and the variety among those faces using OpenFace's Comparison method \cite{amos2016openface} and \textit{face\_recognition's} compare method \cite{face-recognition}.  Our implementation of StyleGAN2 is trained on  a  dataset of 132,000 images taken from stills of YouTube videos of TV night show host John Oliver and from this we synthesise 1,000 images in a way that includes a lot of facial variation. We then use these 1,000 images to train deepfacelab \cite{petrov2020deepfacelab} to generate a deepfake where the (synthesised) John Oliver is superimposed on a subject shown interacting with a chatbot in a dialogue.  A schematic of the flow of our data processing is shown in Figure~\ref{fig:System}.

As we show later in this paper, when trained with enough facial variations in input images, we found that deepfacelab is able to produce an accepted quality of generated deepfakes.

\section{Evaluation Metrics}
\label{sec:evals}

There are a number of methods developed to evaluate the quality of output produced by GANs and to measure the variability in a set of images of faces and we discuss some of these here. For a more detailed description of GAN output see \cite{borji2019pros}.

\subsection{Inception Score (IS)}
Inception Score was first introduced by  Salimans \textit{et} \textit{al.} \cite{salimans2016improved}, and is the most common method used for evaluating GAN outputs. It uses a pre-trained inception model to classify generated images and  calculates probabilities of each image belonging to each class, and looks at the label distribution. Images with high probability towards one class/label are considered high quality. 

In summary, Inception Score actually captures two properties of a generated dataset:
\begin{enumerate}
    \item Image Quality: How highly an image belongs to one class as classified by an inception classifier \ldots do they look similar to a specific object?
    \item Image Diversity: How many different images are generated by the GAN \ldots is there a range of different objects generated?
\end{enumerate}

\noindent 
Inception score has a lowest value of 1.0 and higher values indicate an improving quality of the GAN \cite{inception-mastery}.
However, even with these properties, IS has its limitations as shown in \cite{borji2019pros}.
Firstly, it favours GANs which can store training data and generate images around centers of data modes and secondly, since this method uses an Inception Classifier which is trained on the ImageNet dataset with many object classes, it may uplift those models which produce good images of objects.
A third limitation of IS is that since the score never takes a real dataset into account and evaluates the quality of a GAN based on it's generated dataset, this can be deceptive. This may favour  GANs which produce clear and diverse images of any object, far from a real dataset.

\subsection{Fr{\'e}chet Inception Distance (FID)}

FID is another popular method for GAN evaluation introduced by Heusel \textit{et al.} in 2017 \cite{heusel2017gans}. It uses feature vectors of real data and generated data and calculates distances between them. The FID score is used to evaluate the quality of images generated by GANs, and lower scores have been shown to correlate well with higher quality generated images \cite{fid-mastery}.

Unlike Inception Score (IS), FID captures the statistics of generated data and compares it with the statistics of real data. It is similar to IS in the way that it also uses the inception v3 model. Instead of  using the last output layer of the model, it uses the last coding layer to capture specific features of the input data. These  are collected for both real and generated data. The distance between two distributions, real and generated, is then calculated using Fr\'{e}chet-distance \cite{alt1995computing} which itself uses  the  Wasserstein-2 distance  which  is a calculation between multi-variate Gaussians fitted to data embedded into a feature space \cite{borji2019pros}.
%
Lower distance values convey that the generated dataset is of high quality and similar to real dataset \cite{fid-mastery}.

A model that generates only one image per class will have a bad FID score whereas the same case will have high IS. FID compares data between real and generated data sets whereas IS only measures diversity and quality of a generated dataset. Unlike IS, data scores will be bad on an FID scale in cases where there is  noise or other additions to the data \cite{borji2019pros}.

\subsection{OpenFace Python Library}
\label{sec:openface}

OpenFace is an open source general-purpose library for face recognition  \cite{amos2016openface} with various features including dlib's face landmark detector \cite{king2009dlib}. Landmarks are  used to crop images  to ensure only facial data is passed to the neural network for training, producing a low-dimensional face representation for the faces in images \cite{amos2016openface}.
OpenFace   includes a function to calculate the squared L2 distance \cite{l2}
between  facial representations, providing a comparison function among faces in a  a dataset.
An image in the dataset can be  paired with every other image in the dataset and the squared L2 distance  computed, ranging  from 0 to 4, with  0 meaning the faces in two compared images are more likely to be of the same person \cite{openface-web}.

In our work we applied OpenFace to an image set generated by StyleGAN2 to measure the degree of variability among the generated faces and we computed the mean and variance of inter-image scores among the images.  To confirm our approach, two datasets of facial images were generated, each with 100 images of the same person taken from a smartphone in burst mode. In one  dataset, the facial expressions were kept the same  and we called this dataset the ``Monotone" dataset. In the second dataset,  various facial expressions were captured  called  the ``Varied" dataset. The number of comparisons this requires is 4,950  for each dataset from which we compute  mean and variance.

\begin{figure}
    \centering
    \includegraphics[width=0.49\columnwidth]{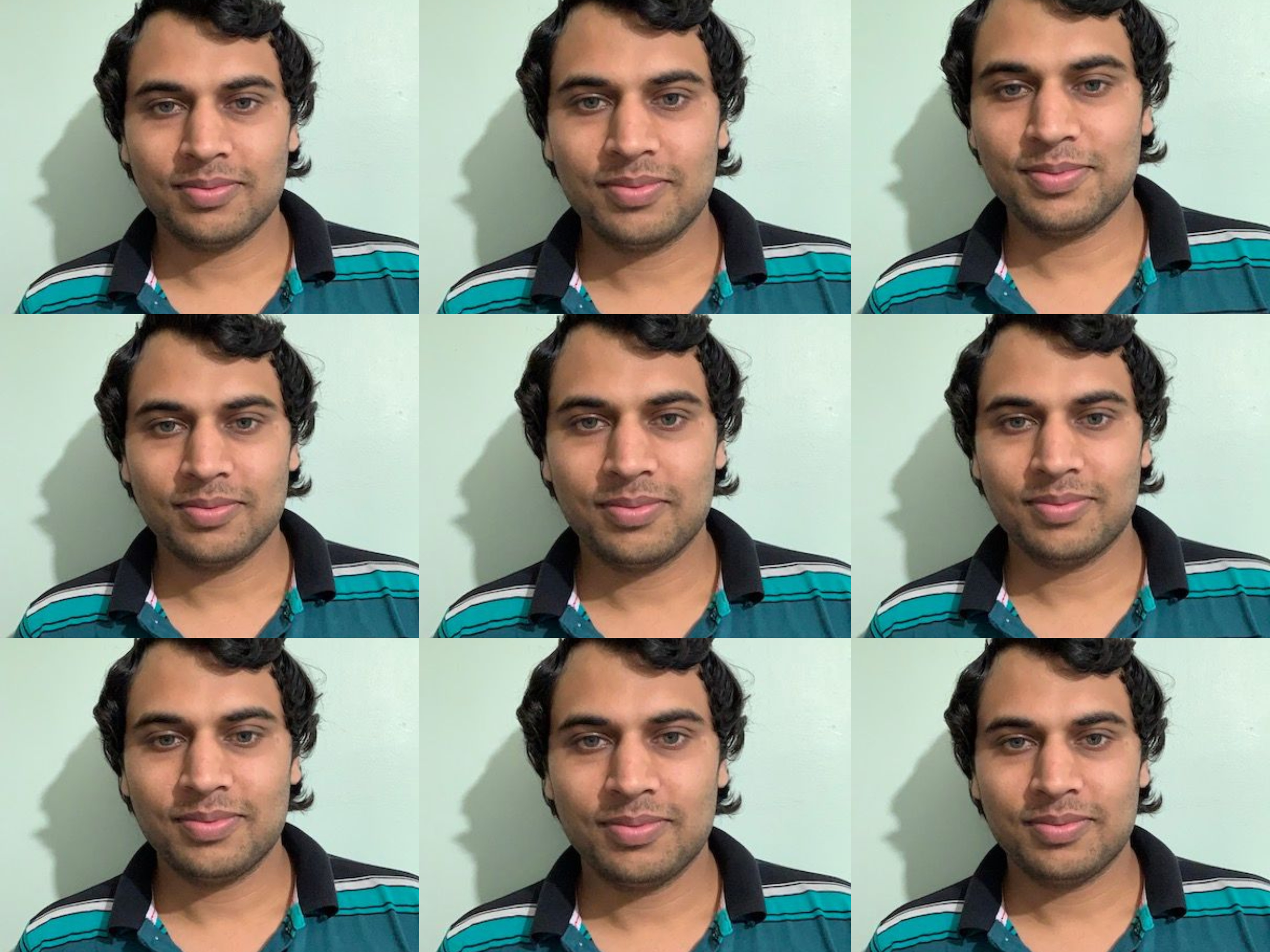}
    \includegraphics[width=0.49\columnwidth]{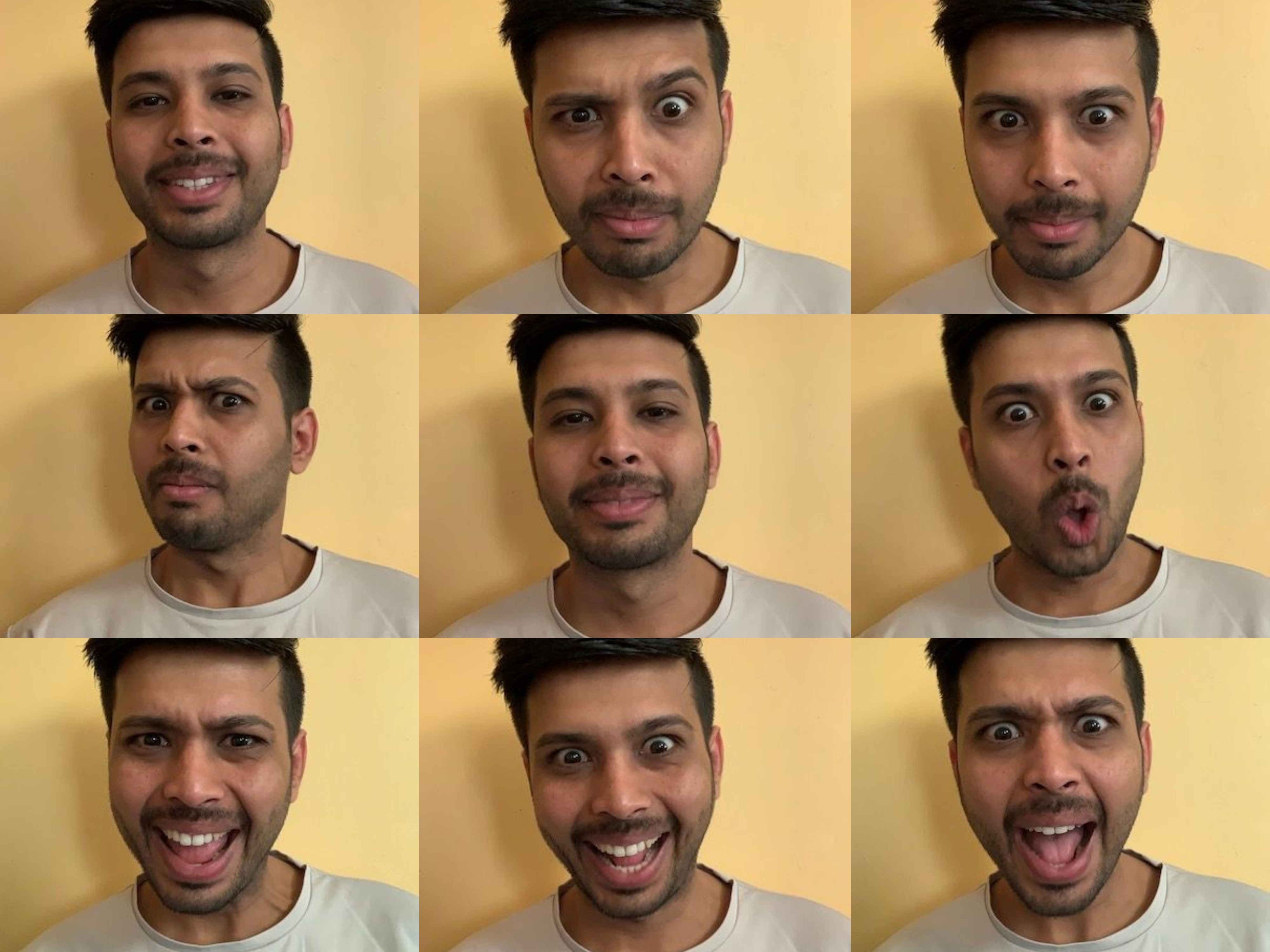}
    \caption{Subset of  Monotone and Varied facial image datasets}
    \label{fig:faces}
\end{figure}

Figure~\ref{fig:faces}  shows a subset of each dataset with calculated mean and variance in Table~\ref{tab:faces}. The Monotone dataset gave a smaller mean and variance score which denotes the person in the dataset is same but with less variation in  facial expression compared to the other dataset which has variability in facial expressions of the individual, though since the mean is still close to zero, the person in the dataset is the same person.

\begin{table}[ht]
    \centering
    \begin{tabular}{lll}
    \toprule
         Dataset&Mean~~&Variance~~  \\
         \midrule
         Monotone faces~~~~~~ & 0.019~~~~~~ & 0.000 \\
         Varied faces & 0.234& 0.019 \\ 
         \bottomrule
    \end{tabular}
    \caption{Mean and variance among two 100 facial image datasets}
    \label{tab:faces}
\end{table}

\subsection{The face\_recognition Python Library}

\begin{figure}[ht]
    \centering
    \includegraphics[width=0.4\columnwidth]{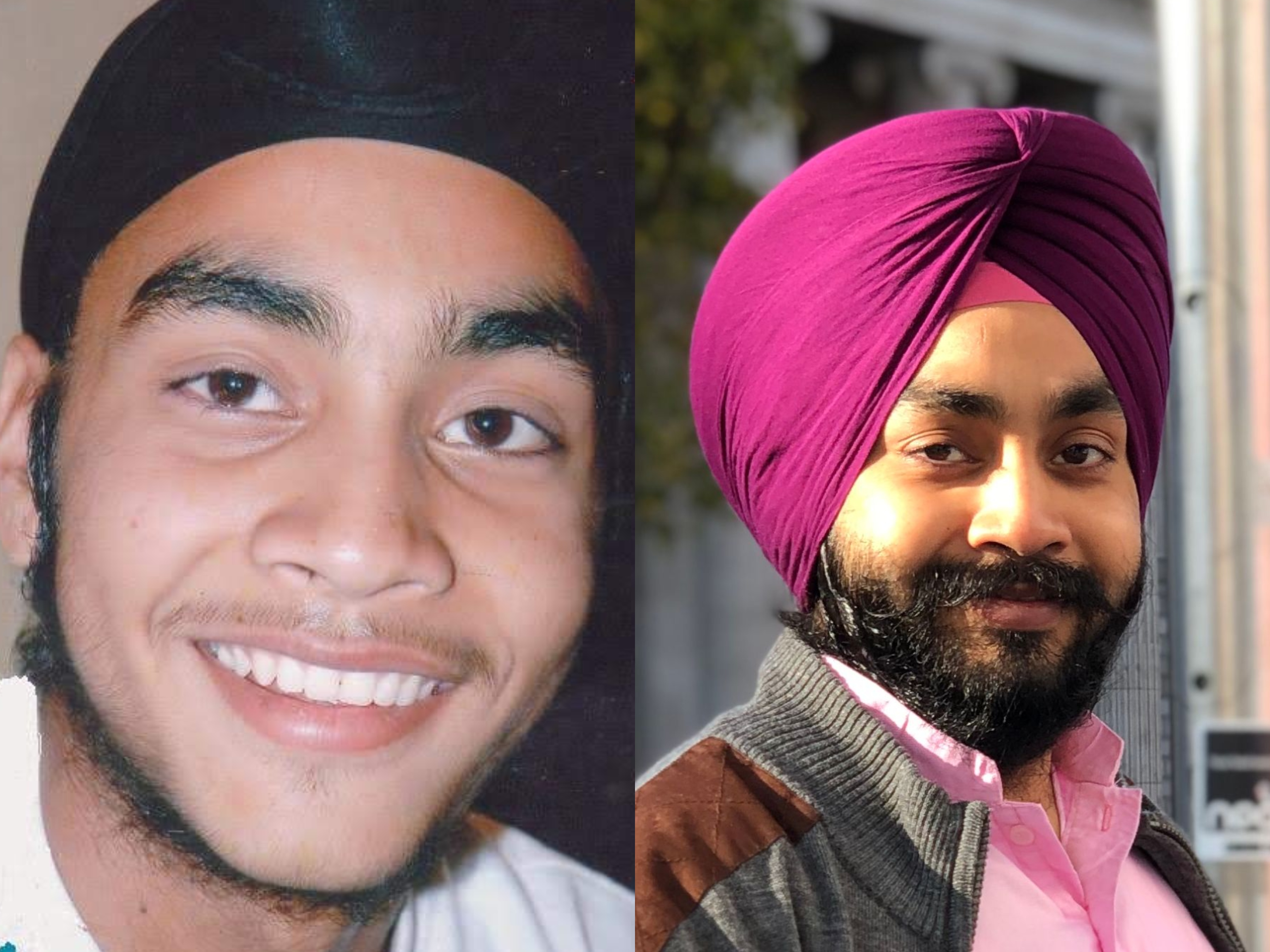}
    \caption{Two images of the same individual taken 12 years apart.}
    \label{fig:age}
\end{figure}

{\it face\_recognition}  is a simple library in Python for face recognition which also uses dlib's facial landmark detector \cite{king2009dlib} and has a comparison feature which calculates distance between facial landmarks of two images. Given a certain threshold, it returns a True/False whether the person in both images is the same or not \cite{face-recognition}.
To show its capabilities power, 
we compared two images of the same individual shown in Figure~\ref{fig:age}, the first taken in 2007 and the second in 2019 and \cite{face-recognition} detects these as the same person. 

\noindent 
For our purposes  we  iterate through the GAN-generated images and compare each with the original images used to train StyleGAN2 using {\it face\_recognition} 
as another way of evaluating the GAN-generated dataset.

\begin{figure*}[ht]
    \centering
    \includegraphics[width=\textwidth]{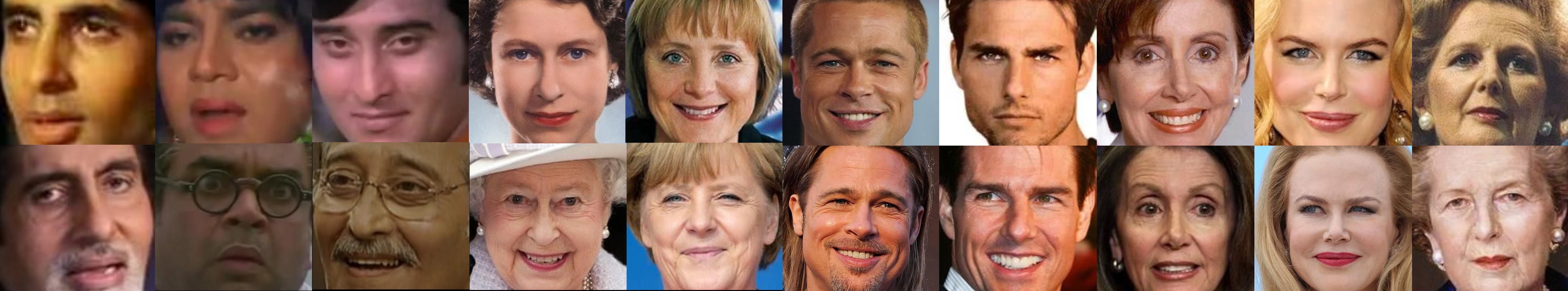}
    \caption{10 images of some celebrities taken years apart. The first row are  images from earlier in their careers and the second row are more recent images.}
    \label{fig:collage}
\end{figure*}

\noindent
To  further validate this method, we took 10 pairs of celebrity face images shown in Figure~\ref{fig:collage}, each pair of images taken years apart \cite{cfp-paper} and using {\it face\_recognition} we compared them, observing that each pair is identified by {\it face\_recognition} as of same person.

\section{Facial Image Data Gathering for GAN Training}

For training a deepfake video generation system there are numerous datasets available from other studies \cite{facial-db} where  facial data was gathered but for almost all of these  the  data was either,
not in sufficient quantity or quality to train a GAN, or consists of faces of different individuals whereas we require images of the same person.

The GAN we  use  is StyleGAN2 developed by  Karras \textit{et al.} in 2019 with improvements over its predecessor StyleGAN \cite{karras2019analyzing} \cite{karras2018stylebased}. StyleGAN2 can generate images up to 1024x1024 pixels in size but this requires  hardware intensive training. We worked at 256x256 pixels image resolution considering the limited hardware  available for this study 
and we generated our own dataset by extracting frames from videos of an individual.

As stated in \cite{john-oliver}, John  Oliver is ``an English-American comedian, writer, producer, political commentator, actor, and television host''. He is the host of a popular HBO Series ``Last week tonight with John Oliver" \cite{john-oliver}. We chose to use videos of him because he is always in the frame and at same position on screen and talks with various facial expressions. His recent videos have a plain background because of being shot at a home studio due to COVID-19.

Using 20  videos from the official YouTube channel\footnote{\url{https://www.youtube.com/user/LastWeekTonight}} we extracted 132,000 frames  cropped to John Oliver's face area with the remaining   part of the frames  ignored. We re-sized the images to 256x256 pixels for model training, using the Pillow python library \cite{pillow}. We   trained StyleGAN2 \cite{karras2019analyzing} by converting to TFRecords format \cite{tfrecord} which took around 30 minutes of processing and around 27GB of storage on a system with 30GB of memory and 1 NVIDIA Tesla V100 GPU on the Google Cloud Platform. 

StyleGAN2 training uses the TFRecord format  \cite{tfrecord} for storing a sequence of binary records which is advantageous in cases of large datasets which cannot be held in memory during training and only require the dataset one at a time (e.g. a batch) when they loaded from disk and processed \cite{style-git}.
The original dataset used by StyleGAN2 training had 70K images from the Flickr Faces HQ (FFHQ) \cite{karras2019analyzing}  at 1024x1024 repeated  25M times. That means if the dataset contains 70K images, the GAN would go over these  repetitively  $25M/70K \simeq 357$  to learn the salient features from 25M images. The authors state  they performed training with 8 GPUs  for almost 10 days to generate high quality images  \cite{style-git}.
%
%
%

\begin{figure}[ht]
    \centering
    \includegraphics[width=0.7\columnwidth]{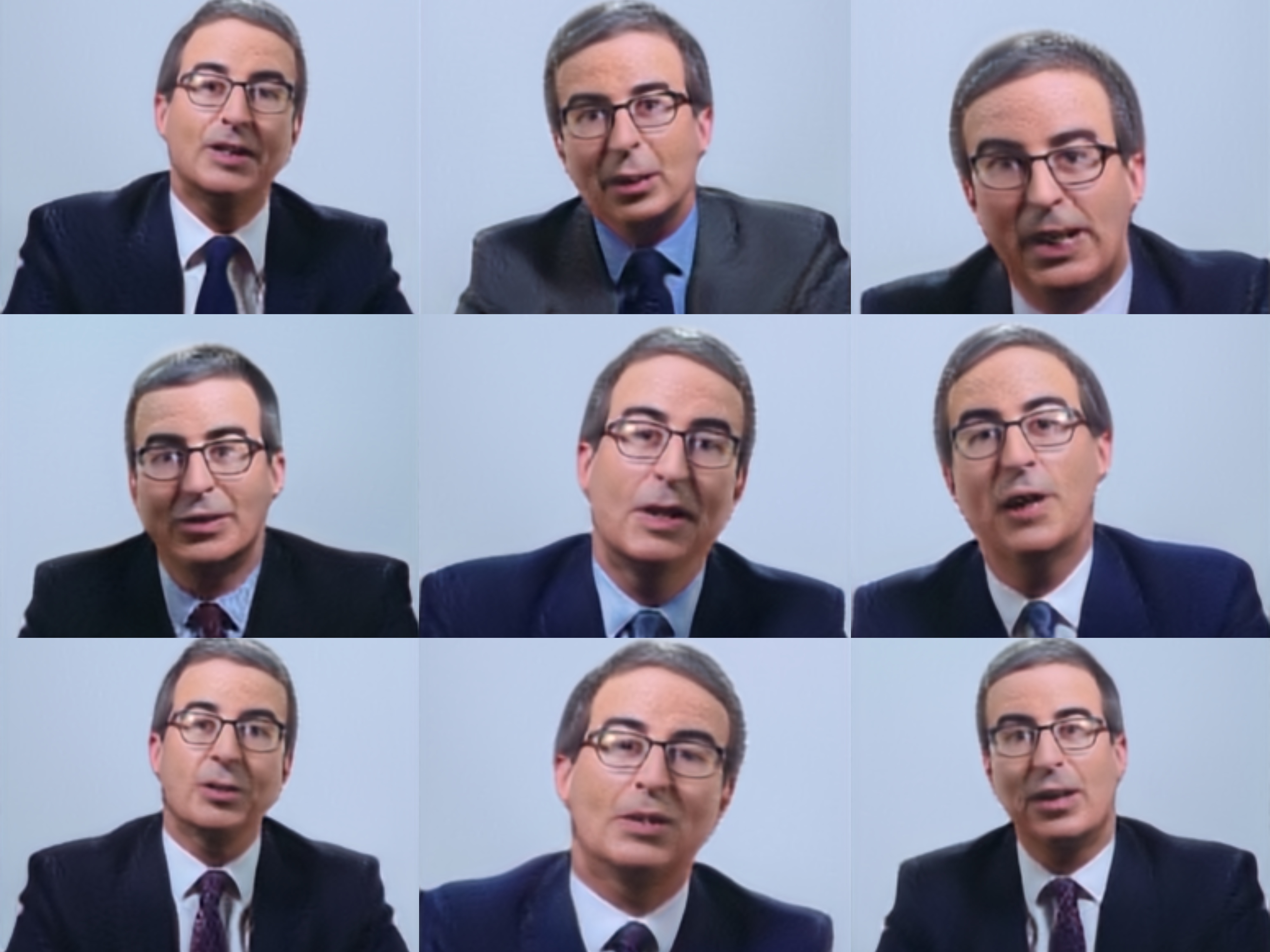}
    \caption{Sample of generated images from John Oliver's trained model}
    \label{fig:john-generated}
\end{figure}

Using our John Oliver dataset of 132,000 images,  training was completed with the number of images  set to 500K. This made the GAN go over the dataset only $500K/132K \simeq 3.8$ times. Since the dataset size was significant and had variation in images even though all images are of a single person, the GAN was able to generate  quality output images and Figure~\ref{fig:john-generated} shows some of these images. 


After generating 1,000  images of John Oliver from our trained model, we applied the metrics described in Section~\ref{sec:evals} and the results are shown in Table~\ref{tab:res}.
Out of 1,000 generated faces,  the face\_recognition library recognised 929 faces as John's faces, rejected 66 images and 5 of the 1,000 images were not recognised as faces due to having  noise in them around the facial landmarks.

\begin{table}[ht]
    \centering
    \begin{tabular}{ll}
    \toprule
         ~~~{\bf Metric}~~~&~~~~{\bf Value}~~~  \\
    \midrule 
        \multirow{2}{*}{~~~~Inception Score $\uparrow$ } &~~~~Mean: 1.635 \\
        &~~~~Variance: 0.011 \\
        \midrule
        ~~~~Fr{\'e}chet Inception Distance $\downarrow$ &~~~~29.318 \\
        \midrule
        \multirow{2}{*}{~~~~OpenFace library comparison $\downarrow$ } &~~~~Mean: 0.401 \\
        &~~~~Variance: 0.068 \\
        \midrule
        \multirow{3}{*}{~~~~face\_recognition library comparison~~~~} &~~~~Matching: 929 \\
        &~~~~Not matching: 66 \\
        &~~~~Not faces: 5 \\
    \bottomrule
    \end{tabular}
    \caption{Scores for 1,000 image output from StyleGAN2 trained on  John Oliver dataset, $\uparrow$  means higher is better and $\downarrow$  means lower is better.}
    \label{tab:res}
\end{table}


These figures gives us certainty that the images generated by the GAN are recognised as John Oliver's face. The comparison feature of the OpenFace Library \cite{amos2016openface} gave a mean score of 0.401 and variance of 0.068 which means that the facial data is of the same person and it has sufficient variability. The calculated Inception score \cite{salimans2016improved} for the same dataset was 1.628.


\section{Deepfake Video Generation}

we used an NLP-based chatbot python library Chatterbot \cite{ChatterBot} to generate dialogues which we could use for our source videos. We built a simple bot and trained it on an English language corpus which has various topics of conversation \cite{chatterbot-corpus}. Since this was  built to give dialogue for our videos, we generated only a few dialogues with the bot and one conversation with random queries after training goes as follows:
\\

{\bf Human:} Do you know gossip?
{\bf Bot:} Gregory said I respond to the current line, not with respect to the entire conversation. Does that count as gossip?
{\bf Human:} What is AI?
{\bf Bot:}Artificial Intelligence is the branch of engineering and science devoted to constructing machines that think.
{\bf Human:} What is a super computer?
{\bf Bot:} Computers which can perform very large numbers of calculations at very high speed and accuracy are called super computers.
{\bf Human:} How does a computer work?
{\bf Bot:} Computers are very dumb. They only execute instructions given by humans.
{\bf Human:} What was the first computer?
{\bf Bot:} It's hard to say, but The ENIAC is regarded as the first 'real' computer. It was developed at University of Pennsylvania in 1946.

\begin{figure}[ht]
    \centering
    \includegraphics[width=0.6\columnwidth]{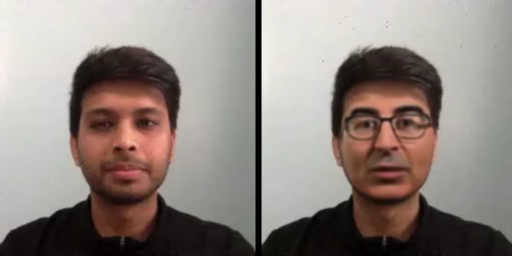}
    \caption{Still frame from video illustrating face swapping (deepfake) from our subject (left side) to John Oliver (right side), video is available at \url{https://bit.ly/31xEjgy}}
    \label{fig:final}
\end{figure}

\noindent 
We recorded a video with our own subject responding in the above dialogue.  The DeepFaceLab \cite{petrov2020deepfacelab} face swapping framework was then used to swap the face of our subject with faces from the 1,000 GAN-generated facial images of John Oliver. DeepFaceLab's model was trained to find the facial area of our subject in our recorded video while the GAN-generated John Oliver facial images were fed to the model. The model was trained for almost 80,000 iterations to learn the features from our subject's face and John Oliver's face and swap them. The output of this swap was a regenerated video dialogue at 256x256 pixel resolution with John Oliver's face  having same dialogue as our subject, in other words it is a deepfake. A sill image is shown in Figure~\ref{fig:final} and the video is available for anonymous viewing at \url{https://bit.ly/31xEjgy}

\section{Conclusions and Future Work}
In this study we introduced and developed an idea to consolidate various techniques available to develop a video dialogue of an individual speaking to camera based on a limited dataset of images  of the individual. We generated a dataset of 132,000 video frames extracted  from TV night show host John Oliver's YouTube videos and trained the StyleGAN2 \cite{karras2019analyzing} GAN  to generate a sample of 1,000 images and  4 evaluation methods were used to measure the variability and quality  of these images. 
These included the Python libraries OpenFace and face\_recognition which  measure facial variability in a dataset of faces.

We then generated several dialogues from a chatbot we trained and  recorded a video with our own subject responding as part of one of these dialogues. We applied a Face Swapping Framework DeepFaceLab \cite{petrov2020deepfacelab} to swap the face of our subject with that of the GAN-generated John Oliver images. The final video output of swapped dialogues alongside the original dialogues is publicly and anonymously available at \url{https://bit.ly/31xEjgy}.

We observe that the deepfake video based on a synthetic set of 1,000 images of  John Oliver is of good quality. There is some colour variation across frames which we could easily have smoothed using a tool like OpenCV but we decided to leave it there to emphasise to the viewer how the video was created. 

Our future work is to repeat the video generation process using a more homogeneous set of images generated by the GAN which synthesises images of John Oliver, and then to compare the quality of the generated deepfakes. While most work on deepfakes has been to detect them, such as 
\cite{dolhansky2019deepfake}, there is little work reported to date on measuring their quality so ultimately the measure of deepfake quality may be how easily it is to be recognised as a deepfake.

\subsubsection*{Acknowledgments.} We wish to thank Satyam Ramawat for acting as a test subject for our image generation and AS is part-funded by Science Foundation Ireland under grant number SFI/12/RC/2289\_P2, co-funded by the European Regional Development Fund.


\end{document}